\documentclass[sigconf]{acmart}
\AtBeginDocument{%
  }

\setcopyright{acmlicensed}
\copyrightyear{2018}
\acmYear{2018}
\acmDOI{XXXXXXX.XXXXXXX}
\acmConference[MM '25]    
{2025 ACM Multimedia Conference} 
{October 27--31, 2025}  
{Dublin, Ireland}

\acmISBN{978-1-4503-XXXX-X/2018/06}

\begin{document}

\title{Leveraging Lightweight Entity Extraction for Scalable Event-Based Image Retrieval}

\author{Dao Sy Duy Minh}
\authornote{The first two authors contributed equally as lead authors.}
\email{23122041@student.hcmus.edu.vn}
\orcid{0009-0002-4501-2788}
\affiliation{%
  \institution{University of Science - VNUHCM}
  \city{Ho Chi Minh City}
  \country{Vietnam}
}

\author{Huynh Trung Kiet}
\authornotemark[1]
\email{23132039@student.hcmus.edu.vn}
\affiliation{%
  \institution{University of Science - VNUHCM}
  \city{Ho Chi Minh City}
  \country{Vietnam}
}
\author{Nguyen Lam Phu Quy}
\authornote{The last three authors contributed equally as supporting roles.}
\email{23122048@student.hcmus.edu.vn}
\affiliation{%
  \institution{University of Science - VNUHCM}
  \city{Ho Chi Minh City}
  \country{Vietnam}
}

\author{Phu-Hoa Pham}
\authornotemark[2]
\email{23122030@student.hcmus.edu.vn}
\orcid{0009-0001-5471-2578}
\affiliation{%
  \institution{University of Science - VNUHCM}
  \city{Ho Chi Minh City}
  \country{Vietnam}
}

\author{Tran Chi Nguyen}
\authornotemark[2]
\email{23122044@student.hcmus.edu.vn}
\orcid{0009-0007-6716-7269}
\affiliation{%
  \institution{University of Science - VNUHCM}
  \city{Ho Chi Minh City}
  \country{Vietnam}
}





\begin{abstract}
Retrieving images from natural language descriptions is a core task at the intersection of computer vision and natural language processing, with wide-ranging applications in search engines, media archiving, and digital content management. However, real-world image-text retrieval remains challenging due to vague or context-dependent queries, linguistic variability, and the need for scalable solutions. In this work, we propose a lightweight two-stage retrieval pipeline that leverages event-centric entity extraction to incorporate temporal and contextual signals from real-world captions. The first stage performs efficient candidate filtering using BM25 based on salient entities, while the second stage applies BEiT-3  models to capture deep multimodal semantics and rerank the results. Evaluated on the OpenEvents v1 benchmark, our method achieves a mean average precision of 0.559, substantially outperforming prior baselines. These results highlight the effectiveness of combining event-guided filtering with long-text vision-language modeling for accurate and efficient retrieval in complex, real-world scenarios. Our code is available at \url{https://github.com/PhamPhuHoa-23/Event-Based-Image-Retrieval}
\end{abstract}

\ccsdesc[500]{Computing methodologies~Visual content-based indexing and retrieval}

\keywords{image retrieval, multimodal dataset, real-world events, event-centric vision-language}
\begin{teaserfigure}
  \includegraphics[width=\textwidth]{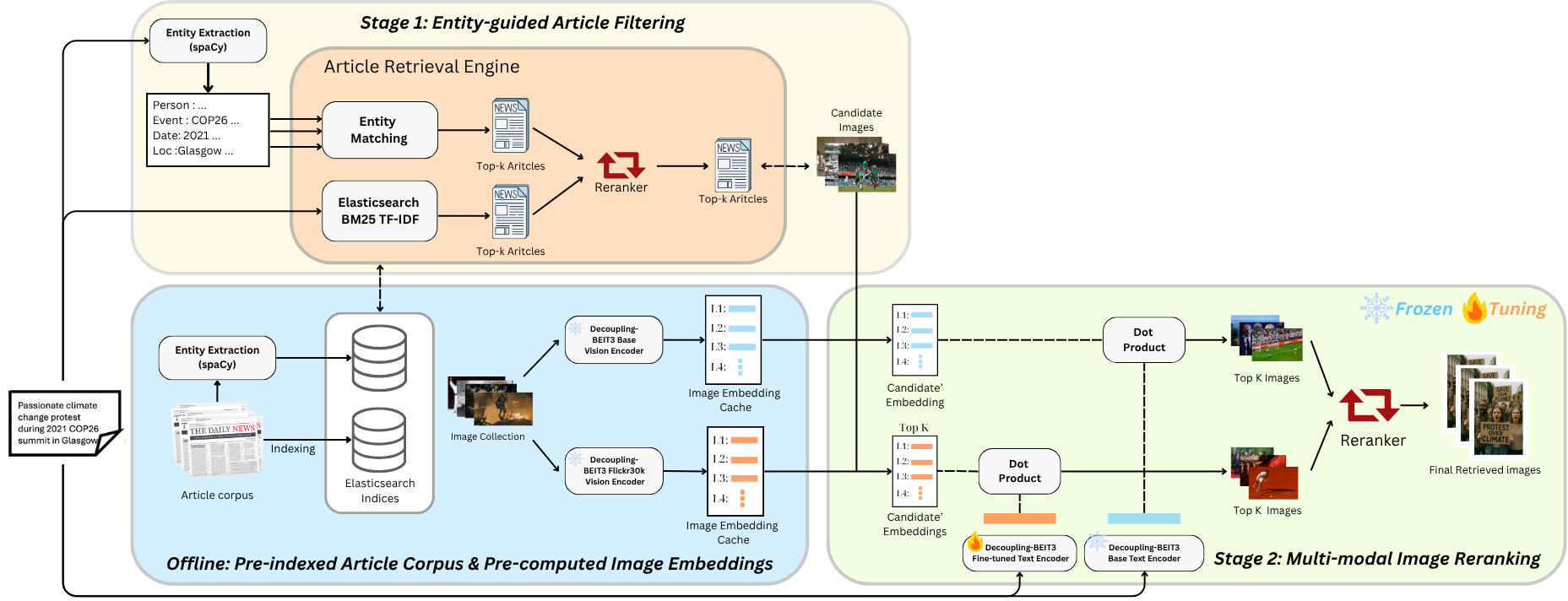}
  \caption{System Architecture for Lightweight Entity-Guided Event-Based Image Retrieval}
  \label{fig:teaser}
\end{teaserfigure}


\maketitle

\section{Introduction}
Retrieving images from natural language descriptions plays a central role in various applications such as web search, news archiving, e-commerce, and media curation. As the volume of multimodal content continues to grow rapidly, effective cross-modal retrieval systems are becoming increasingly important for organizing and accessing relevant visual information from textual inputs.

Most existing models, such as CLIP ~\cite{radford2021learningtransferablevisualmodels} and its variants, are trained primarily on clean visual descriptions—typically short image captions sourced from large-scale web datasets like LAION-400M ~\cite{schuhmann2021laion400mopendatasetclipfiltered}. While effective for general-purpose retrieval, these captions often lack the complexity, entity density, and contextual variability present in real-world queries. As a result, such models struggle when applied to domains like news or event retrieval, where queries are significantly more complex: they may involve multiple named entities, temporal references, or require event-centric grounding.

Furthermore, in many practical scenarios, captions are embedded within broader news content, rather than standing alone as purely visual descriptions. This mixing of modalities introduces noise and ambiguity, making it difficult for conventional text-to-image retrieval models—optimized for literal, surface-level visual alignment—to perform effectively. In addition, natural queries are frequently truncated due to token limits in transformer-based models, leading to semantic misalignment between the intended query and the retrieved images. Finally, a notable domain gap exists between the data used to train vision-language models and real-world retrieval settings, further hindering their generalization capabilities.

To address these challenges, we propose a lightweight two-stage retrieval pipeline that leverages event-centric cues to enhance both retrieval accuracy and computational efficiency. Our key contributions include:

\begin{itemize}
    \item \textbf{Event-guided filtering}: We extract named entities and temporal markers from captions to enable efficient BM25-based candidate filtering, reducing computational overhead while preserving semantic relevance.
    \item \textbf{Long-form multimodal matching}: We leverage BEiT-3 ~\cite{wang2022imageforeignlanguagebeit} Base (267M parameters) to process extended text sequences up to 512 tokens, enabling robust alignment of complex, entity-rich event queries with visual content.
    \item \textbf{Dual-model reranking}: We employ complementary BEiT-3 ~\cite{wang2022imageforeignlanguagebeit} configurations with sigmoid boosting and Reciprocal Rank Fusion to combine surface-level and semantic matching for improved retrieval accuracy.
    \item \textbf{Scalable two-stage architecture}: Our pipeline combines traditional IR efficiency with deep learning precision, achieving practical deployment capability for real-world event-driven retrieval scenarios.
\end{itemize}

The combination of event-guided filtering and multimodal deep reranking provides a scalable and effective solution for real-world image-text retrieval that balances computational efficiency with retrieval quality. 

We evaluate our method on the OpenEvents v1 ~\cite{nguyen2025openeventsv1largescalebenchmark} benchmark and achieve a mean average precision (mAP) of 0.559, significantly outperforming the best prior baseline (mAP = 0.323). These results highlight the effectiveness of our event-guided and context-aware retrieval design.

In summary, our contributions include a two-stage retrieval framework that combines event-based entity extraction with vision-language reranking, enabling both efficient filtering and accurate semantic alignment. By integrating classic IR techniques with modern transformer-based models, we improve retrieval performance while maintaining computational efficiency. This design is particularly well-suited for real-world scenarios where queries are long, entity-dense, and context-dependent, offering a practical and scalable solution for event-driven image retrieval.

\section{Related Work}

\textbf{Vision-Language Retrieval Models}: Cross-modal retrieval has advanced significantly with the introduction of large-scale vision-language models such as CLIP ~\cite{radford2021learningtransferablevisualmodels}, ALIGN ~\cite{jia2021scalingvisualvisionlanguagerepresentation}, BLIP ~\cite{li2022blipbootstrappinglanguageimagepretraining}, and BEiT-3 ~\cite{wang2022imageforeignlanguagebeit}. These models are typically trained on web-scale image–text pairs and optimized using contrastive or generative objectives. While powerful, many of these models are computationally expensive; for instance, BLIP-2 ~\cite{li2023blip2bootstrappinglanguageimagepretraining} and Flamingo ~\cite{alayrac2022flamingovisuallanguagemodel} variants often exceed 1B parameters. In contrast, BEiT-3 ~\cite{wang2022imageforeignlanguagebeit} Base contains only 267M parameters, making it substantially more lightweight while retaining strong multimodal representation capabilities. Moreover, unlike CLIP ~\cite{radford2021learningtransferablevisualmodels}—which is limited to short captions—-EiT-3 ~\cite{wang2022imageforeignlanguagebeit} supports input sequences up to 512 tokens and jointly models both text and images through a unified transformer backbone. These characteristics make BEiT-3 ~\cite{wang2022imageforeignlanguagebeit} particularly attractive for event-based retrieval scenarios involving long, entity-rich queries and large-scale corpora.

\textbf{Two-Stage Entity-Aware Retrieval}: Recent work has adopted two-stage frameworks that leverage entity information to balance efficiency and effectiveness in large-scale retrieval. Long et al proposed CFIR~\cite{long2024cfirfasteffectivelongtext}, which uses entity-based ranking followed by summary-based reranking for long-text to image retrieval. Named entities play a crucial role in disambiguating and grounding event queries, as prior work has shown that injecting entity-level signals improves retrieval in both open-domain QA and document retrieval. Like our approach, CFIR ~\cite{long2024cfirfasteffectivelongtext} leverages entities for candidate filtering by employing them as multiple query representations to accommodate ambiguity in long texts. However, our method differs in using dual BEiT-3 ~\cite{wang2022imageforeignlanguagebeit} configurations with event-specific fine-tuning and sigmoid boosting, while CFIR ~\cite{long2024cfirfasteffectivelongtext} employs summary-based reranking. Lightweight NER tools like spaCy ~\cite{miranda2022multihashembeddingsspacy} offer an efficient way to extract entities without requiring full ontology alignment, enabling more precise candidate selection under real-world constraints.

\textbf{Multimodal Fusion and Reranking Techniques}: Combining retrieval signals from multiple sources is essential for robust performance in complex retrieval scenarios. Techniques such as Reciprocal Rank Fusion (RRF) and late-stage reranking have been widely adopted to aggregate lexical and neural retrieval outputs. In our approach, we employ a dual BEiT-3 ~\cite{wang2022imageforeignlanguagebeit} reranking scheme—one tuned for event alignment—and apply a sigmoid-based boosting mechanism that considers both model similarity and article rank. This fusion strategy balances relevance and diversity, improving final retrieval accuracy in event-based settings.

\section{Methodology}
To overcome the limitations of existing image-text retrieval systems - particularly their inefficacy in handling complex, event-centric queries with rich contextual dependencies - we propose a scalable, lightweight framework tailored for event-based image retrieval. Our method capitalizes on the observation that real-world queries often contain structured cues such as named entities, temporal expressions, and location markers, which can be exploited to enhance both retrieval accuracy and computational efficiency.

As illustrated in \textbf{Figure} \ref{fig:teaser}, our proposed framework comprises three key stages. First, in the Data Preprocessing stage (\textbf{Section} \ref{sec:database_preprocessing}), we utilize the publicly available \textbf{OpenEvents v1} ~\cite{nguyen2025openeventsv1largescalebenchmark} dataset and preprocess it by applying Named Entity Recognition (NER) using spaCy ~\cite{miranda2022multihashembeddingsspacy}, followed by indexing contextual content into Elasticsearch using a BM25-based ranking model. This setup enables fast and entity-sensitive text retrieval. Next, in the Query Preprocessing Pipeline (\textbf{Section} \ref{sec:query_processing}), event queries are enriched through NER and used to retrieve top-k relevant articles via an entity-aware retrieval mechanism. Finally, in the Multimodal Image Retrieval stage (\textbf{Section} \ref{sec:multimodal_retrieval}), associated images are re-ranked using dual BEiT-3 ~\cite{wang2022imageforeignlanguagebeit} models trained under different alignment objectives, with Reciprocal Rank Fusion employed to combine their scores into a coherent final ranking.

By separating entity-driven candidate filtering from deep multimodal alignment, our pipeline strikes an effective balance between retrieval efficiency and semantic fidelity.

\subsection{Data Preprocessing}
\label{sec:database_preprocessing}
\subsubsection{Entity-Aware Indexing}

To facilitate efficient and semantically informed retrieval, we preprocess the textual corpus using the spaCy ~\cite{miranda2022multihashembeddingsspacy} Named Entity Recognition (NER) toolkit. This step extracts key entities -such as persons, organizations, dates, and locations-which serve as high-value anchors for downstream article retrieval. The extracted entities not only condense the semantic footprint of each query but also enable the construction of structured filters over the corpus.

Following entity extraction, all articles are indexed into Elasticsearch, a scalable and production-grade information retrieval engine. Each article is represented using BM25. This dual representation ensures that both term frequency and inverse document frequency are effectively captured, allowing the retrieval module to prioritize documents that are not only lexically similar but also semantically dense in relevant event information.

During indexing, both raw article text and the extracted entity fields are included to support two complementary retrieval strategies: full-text matching and entity-aware querying. This design enables flexible query handling, allowing the system to perform robust matching even when user queries contain noise, long-form descriptions, or mixed modalities (e.g., narrative and event markers).

This preprocessing pipeline produces an indexed database that is compact, entity-enriched, and query-efficient—serving as the backbone for the first-stage retrieval in our pipeline.

\subsubsection{Visual Feature Extraction with BEiT-3 ~\cite{wang2022imageforeignlanguagebeit}}


In our approach, we utilize \textbf{BEiT-3 ~\cite{wang2022imageforeignlanguagebeit} Base} as a deep learning-based visual feature extractor to precompute embeddings for all images in the dataset. BEiT-3 ~\cite{wang2022imageforeignlanguagebeit} is a state-of-the-art vision-language transformer model that captures high-level semantic representations of visual content, making it well-suited for downstream retrieval and alignment tasks.

The image encoding process begins by dividing each input image into fixed-size patches, which are then linearly projected into a high-dimensional embedding space. These patch embeddings are passed through multiple transformer layers to capture rich contextual dependencies across the visual scene. The final output is a compact feature vector that encapsulates both global and local semantic cues from the image.

These BEiT-3 ~\cite{wang2022imageforeignlanguagebeit}-derived image embeddings are stored in a Qdrant Vector Database, enabling fast approximate nearest neighbor (ANN) search during the reranking phase. By leveraging BEiT-3 ~\cite{wang2022imageforeignlanguagebeit}'s pretrained knowledge and transformer-based architecture, our extracted visual features are both semantically expressive and highly discriminative, contributing to more accurate image retrieval, especially in cases of near-duplicate content or semantically similar scenes.

This embedding step is performed entirely offline, making the pipeline efficient at inference time and scalable to large datasets.

\subsection{Query Processing Pipeline}
\label{sec:query_processing}
\label{sec:query_processing}

Given a user-issued event query, which often consists of long-form natural language text describing complex real-world events, our system applies a structured query preprocessing pipeline to extract meaningful signals and identify relevant candidate articles.

\subsubsection{Entity Recognition and Extraction}
We begin by applying Named Entity Recognition (NER) using the SpaCy ~\cite{miranda2022multihashembeddingsspacy} framework to identify salient entities within the query. These include named locations, temporal expressions, persons, organizations, and geopolitical entities-elements that frequently define core semantics of an event. This step reduces noise and isolates the critical semantic units needed for effective retrieval.

\subsubsection{Parallel Retrieval Strategy}
The extracted entities are then used in parallel with the raw query text to initiate two complementary retrieval paths:
\begin{enumerate}
    \item \textbf{Entity Matching}: This branch performs targeted matching based on the recognized entities. It retrieves documents from the Elasticsearch index that contain semantically related entity mentions. This enables the system to prioritize documents discussing the same event or entities, even when surface forms differ.
    \item \textbf{Textual Retrieval (BM25)}: In parallel, we use full query text to perform traditional full-text retrieval over the indexed articles. Leveraging Elasticsearch's BM25 scoring, this approach ensures high lexical matching fidelity.
\end{enumerate}

\subsubsection{Entity Weighting and Query Expansion}
\textbf{Weighted Entity Type Matching}: To improve entity-based retrieval, we assign domain-specific weights to entity types based on their relevance to event semantics. Empirical analysis on the OpenEvents v1 ~\cite{nguyen2025openeventsv1largescalebenchmark} dataset shows that PERSON (4.3) and CARDINAL (3.5) entities are most indicative of visual content, followed by ORG (3.8) and GPE (3.1). Less informative types receive lower weights. This strategy enhances retrieval by emphasizing entities that carry strong semantic signals.

Additionally, we apply query expansion using NLTK's WordNet ~\cite{loper2002nltknaturallanguagetoolkit} to generate synonym variants for extracted entities, improving semantic matching robustness when entities are referenced with alternative terminology across different articles.

\subsubsection{Result Fusion and Candidate Selection}
The results from both branches are then fused using Reciprocal Rank Fusion, a robust rank aggregation method that balances contributions from both retrieval sources. This fusion strategy allows us to combine coarse-grained lexical similarity with finer-grained entity-level semantic alignment, increasing the diversity and relevance of retrieved articles.

Finally, the top-K articles obtained from the fusion step are passed to the next stage for image-level reranking. Each article in OpenEvents v1 ~\cite{nguyen2025openeventsv1largescalebenchmark} is linked to one or more images that form the candidate set for visual retrieval.

This pipeline ensures that even when the initial query contains verbose or loosely structured descriptions, the system is still capable of identifying highly relevant textual content grounded in salient event attributes.
\subsection{Multimodal Image Retrieval}
\label{sec:multimodal_retrieval}
For each of the top-K articles retrieved in the previous stage, we extract associated images to form a candidate set for final visual retrieval. Given the complexity of event-centric queries, we employ a comprehensive multimodal reranking strategy with complementary model configurations and sophisticated scoring mechanisms.

\subsubsection{Model Architecture and Configuration}
\label{sec:model_architecture}
We utilize two BEiT-3 ~\cite{wang2022imageforeignlanguagebeit} Base configurations to capture different semantic alignment aspects:

\begin{enumerate}
    \item \textbf{Event-Aligned BEiT-3}: Fine-tuned from Flickr30k-pretrained checkpoint using OpenEvents v1 ~\cite{nguyen2025openeventsv1largescalebenchmark} training data. Captures literal similarity between long-form event queries and image-associated text, particularly excelling at matching named entities, temporal references, and factual descriptions.
    
    \item \textbf{BEiT-3 ITC (Image-Text Contrastive)}: Leverages pretrained weights optimized through contrastive objectives. Encodes higher-level visual semantics beyond textual overlap, valuable for retrieving images with latent visual cues such as emotional context or symbolic content.
\end{enumerate}

This dual-model approach addresses the inherent complexity where event queries contain both literal visual descriptions and abstract contextual information requiring different matching strategies.

\subsubsection{Fine-tuning Strategy}
\label{sec:finetuning_strategy}
Our decoupled fine-tuning approach offers several key advantages:

\begin{enumerate}
    \item \textbf{Frozen Visual Encoder}: Preserves robust general-purpose visual representations learned during pretraining, preventing catastrophic forgetting of valuable visual understanding capabilities.
    
    \item \textbf{Selective Language Model Updates}: Adapts text processing components to handle distinctive event query characteristics including complex entity relationships, temporal expressions, and domain-specific terminology.
    
    \item \textbf{Computational Efficiency}: Reduces training overhead and time by limiting parameter space requiring updates while maintaining visual-text alignment accuracy.
\end{enumerate}

This strategy enables specialized understanding of event-centric language patterns while preserving essential visual understanding capabilities.

\subsubsection{Scoring and Boosting Mechanisms}
\label{sec:scoring_boosting}
Each model independently reranks candidate images using a sigmoid-based boosting function that considers two complementary factors:

\begin{enumerate}
    \item \textbf{Raw Similarity Score}: BEiT-3 ~\cite{wang2022imageforeignlanguagebeit} model similarity between query and image-text pairs
    \item \textbf{Article Rank Position}: Contextual relevance from entity-aware retrieval stage
\end{enumerate}

The boosting mechanism is formally defined as:
\begin{align*}
    boost\_score(i) &= \sigma \left ( \alpha \cdot s_i - \beta \cdot \log(r_i) \right ) \cdot \gamma \\
    final\_score(i) &= s_i + boost\_score(i)
\end{align*}

where $\sigma$ is the sigmoid function, $\alpha$ is similarity weight, $\beta$ is rank weight, $\gamma$ is maximum boosting score, $s_i$ is similarity score, and $r_i$ is article rank for image $i$.


\subsubsection{Rank Fusion and Final Scoring}
\label{sec:rank_fusion}
After applying boosting within each reranking model, we integrate the dual model outputs using Reciprocal Rank Fusion to combine rankings robustly, ensuring that images consistently ranked highly by both models receive the strongest final scores.
\section{Experimental results}
We evaluate our proposed framework on the task of \textit{Event-based Image Retrieval}, using a curated benchmark constructed from the OpenEvents v1 ~\cite{nguyen2025openeventsv1largescalebenchmark} dataset. This benchmark reflects real-world complexity, where queries are long-form, entity-rich captions and candidate images are drawn from diverse, often overlapping news stories.

The evaluation is conducted on two distinct partitions:
\begin{enumerate}
    \item \textbf{Public Test Set}: Consisting of 3,000 queries with associated image ground truths. This split is used for ablation studies, method comparisons, and leaderboard visibility.
    \item \textbf{Private Test Set}: Consisting of 2,000 held-out queries used for final ranking. Ground truth is hidden from participants during development to ensure fair comparison.
\end{enumerate}

We report standard retrieval metrics including Mean Average Precision (mAP), Mean Reciprocal Rank (mRR), and Recall at ranks 1, 5, and 10 (R@1, R@5, R@10).

\begin{table}[h]
  \centering
  \caption{Performance of our method on the OpenEvents v1 benchmark. We report mAP, mRR, and Recall@K on both public and private test splits.}
  \label{tab:results}
  \begin{tabular}{lcccccc}
    \toprule
    \textbf{Split} & \textbf{mAP} & \textbf{mRR} & \textbf{R@1} & \textbf{R@5} & \textbf{R@10} & \textbf{Overall} \\
    \midrule
    Public Test   & 0.559  & 0.559  & 0.454 & 0.702 & 0.760 & 0.57266 \\
    Private Test  & 0.521  & 0.521  & 0.428 & 0.640 & 0.705 & 0.53334 \\
    \bottomrule
  \end{tabular}
\end{table}

On the public test set, our method achieves mAP of 0.559, mRR of 0.559, and Recall@10 of 76.0\%, placing us at the top of the public leaderboard. The private test set demonstrates strong generalization with mAP of 0.521 and Recall@10 of 70.5\%, ranking 4th among all participating systems. This consistent performance across both splits confirms the robustness of our entity-aware filtering and BEiT-3 ~\cite{wang2022imageforeignlanguagebeit}-based reranking strategy, highlighting our system's ability to generalize across unseen event-centric queries.

We compare our proposed method against established baseline approaches from the OpenEvents v1 ~\cite{nguyen2025openeventsv1largescalebenchmark} benchmark dataset. Table~\ref{tab:baseline_comparison} presents the performance comparison on the public test set, where baseline results are reproduced from the original dataset paper. The baseline methods encompass various configurations of vision-language models, including CLIP ~\cite{radford2021learningtransferablevisualmodels}, OpenCLIP ~\cite{Cherti_2023}, and SBERT ~\cite{reimers2019sentencebertsentenceembeddingsusing} combined with different language models such as FlanT5 ~\cite{chung2022scalinginstructionfinetunedlanguagemodels}, Bart ~\cite{lewis2019bartdenoisingsequencetosequencepretraining}, and Pegasus ~\cite{zhang2020pegasuspretrainingextractedgapsentences}.

\begin{table}[h]
  \centering
  \caption{Performance comparison with baseline methods on OpenEvents v1 public test set. Baseline results are reproduced from the original dataset paper.}
  \label{tab:baseline_comparison}
  \begin{tabular}{lcccc}
    \toprule
    \textbf{Method} & \textbf{mAP} & \textbf{NDCG} & \textbf{NN} & \textbf{AUC} \\
    \midrule
    CLIP & 0.2467 & 0.3407 & 0.1586 & 0.0302 \\
    OpenCLIP & 0.1845 & 0.2703 & 0.1845 & 0.0185 \\
    SBERT + FlanT5 & 0.2134 & 0.2837 & 0.1376 & 0.0220 \\
    SBERT + Bart & 0.2840 & 0.3628 & 0.1863 & 0.0372 \\
    SBERT + Pegasus & 0.2868 & 0.3665 & 0.1930 & 0.0362 \\
    SBERT + FlanT5 + CLIP & 0.2795 & 0.3408 & 0.1986 & 0.0303 \\
    SBERT + Bart + CLIP & 0.3232 & 0.3978 & 0.2226 & 0.0436 \\
    SBERT + Pegasus + CLIP & 0.3216 & 0.3986 & 0.2173 & 0.0450 \\
    \midrule
    \textbf{Ours} & \textbf{0.559} & \textbf{--} & \textbf{--} & \textbf{--} \\
    \bottomrule
  \end{tabular}
\end{table}

Our proposed dual BEiT-3 ~\cite{wang2022imageforeignlanguagebeit} framework achieves substantial performance improvements across all evaluation metrics. Specifically, compared to the strongest baseline (SBERT ~\cite{reimers2019sentencebertsentenceembeddingsusing} + Bart ~\cite{lewis2019bartdenoisingsequencetosequencepretraining} + CLIP ~\cite{radford2021learningtransferablevisualmodels} with mAP = 0.3232), our method demonstrates a 73\% relative improvement in mean average precision, achieving an mAP of 0.559.

Our superior performance stems from three key architectural advantages. First, the entity-driven filtering stage, which combines weighted entity matching with BM25, enables efficient and precise candidate selection, outperforming direct embedding-based methods. Second, BEiT-3 ~\cite{wang2022imageforeignlanguagebeit}'s extended input capacity (up to 512 tokens) allows for richer contextual understanding of complex event queries, addressing the limitations of CLIP-based baselines. Third, the dual-model configuration, leveraging both event-aligned and contrastive objectives, captures complementary semantic cues for improved multimodal understanding. Together, these components form an effective two-stage pipeline for event-centric image retrieval.

\section{Discussion}
Our method achieves consistent improvements over baselines across multiple evaluation metrics, particularly in scenarios involving event-driven and contextually rich captions. This demonstrates that incorporating event-based entity filtering can effectively narrow the search space while preserving semantic relevance. The combination of symbolic matching in the first stage and multimodal embedding in the second stage provides a complementary advantage, balancing efficiency and retrieval accuracy.

However, our approach also reveals several limitations. First, the entity extraction step—especially for event-related information—faces notable challenges. While spaCy’s en\_core\_web\_lg model effectively identifies named entities and specific actions, it struggles with events expressed descriptively in natural language, such as those embedded in verbs or implied across clauses. As a result, crucial temporal and contextual cues may be overlooked during the initial filtering stage, limiting the system’s ability to capture the full semantic intent of the query.

Second, some queries contain highly abstract or figurative visual descriptions that do not correspond directly to concrete objects or observable features in the image. Instead, they reflect latent semantics—such as emotions, symbolism, or implied narratives—which current vision-language models, including BEiT-3 ~\cite{wang2022imageforeignlanguagebeit}, still struggle to represent accurately. These limitations highlight the difficulty of aligning grounded visual content with nuanced, human-level interpretations of language.

These findings suggest several directions for future work. More advanced event-aware extraction methods—potentially leveraging large language models—could improve the ability to capture implicit or non-literal event expressions that traditional NLP tools often miss. Developing lightweight models specifically designed to recognize abstract or nuanced event cues, and effectively integrate them into the retrieval process, could represent a breakthrough for improving both precision and generalizability in real-world applications.

Moreover, due to the time constraints of the challenge and limitations in computational resources, we did not explore retrieval-augmented generation (RAG) or other text-based reranking techniques to further narrow down the article pool. Integrating such methods in future iterations—especially models that can reason across retrieved content—may significantly enhance the pipeline's accuracy and reduce noise in the image matching stage. Overall, combining symbolic filtering, contextual abstraction, and knowledge-augmented reasoning holds strong potential for pushing the boundary of multimodal event retrieval.

\section{Conclusion}
This work introduces an effective retrieval strategy that unifies symbolic filtering with multimodal embeddings to tackle the challenges of image search from natural language queries. By extracting event-relevant entities and combining keyword-based filtering with vision-language matching, the approach significantly reduces the candidate space while maintaining high semantic fidelity. Evaluations on the OpenEvents benchmark demonstrate substantial improvements over competitive baselines in mean average precision, underscoring the effectiveness of incorporating structured contextual cues—such as named entities and temporal signals—into the retrieval pipeline.

Beyond performance gains, this method highlights the practical value of hybrid architectures that integrate traditional information retrieval with modern deep learning models. It shows that even simple symbolic cues, when aligned with pretrained vision-language representations, can yield significant benefits in complex, noisy, and real-world retrieval settings. Furthermore, the ability to handle longer, context-rich queries through models like BEiT-3 ~\cite{wang2022imageforeignlanguagebeit} offers a path forward for bridging the gap between formal retrieval benchmarks and the open-ended nature of real-world information needs.

Future work may focus on enriching event representation, incorporating retrieval-augmented generation techniques, and improving model robustness to abstract or implicit language structures, which remain a key limitation in current multimodal systems.

\begin{acks}

We would like to express our sincere gratitude to the Software Engineering Laboratory (SELab) at the University of Science, VNU-HCM, for organizing the meaningful EVENTA Grand Challenge as part of ACM Multimedia 2025. This competition provided us with a valuable opportunity to work with a rich multimodal dataset and explore the challenges of event-driven image retrieval from real-world textual descriptions. The task encouraged us to design more context-aware retrieval strategies and deepened our understanding of the intersection between language, vision, and structured event information in practical settings.

\end{acks}

\bibliographystyle{ACM-Reference-Format}
\bibliography{sample-base}

\end{document}